\documentclass[11pt]{article}

\usepackage[preprint]{acl}

\usepackage{times}
\usepackage{latexsym}

\usepackage[T1]{fontenc}

\usepackage[utf8]{inputenc}

\usepackage{microtype}

\usepackage{inconsolata}

\usepackage{graphicx}
\usepackage{amsmath}
\usepackage{amssymb}
\usepackage{booktabs}
\usepackage{subcaption}
\usepackage{multirow}
\usepackage{xcolor}
\usepackage{float}

\usepackage[most]{tcolorbox}
\newcounter{prompt}
\tcbset{promptbox/.style={
  colback=gray!4, colframe=gray!55, boxrule=0.5pt,
  left=4pt, right=4pt, top=4pt, bottom=4pt,
  fonttitle=\footnotesize\bfseries, coltitle=black,
  colbacktitle=gray!18, breakable, enhanced}}

\title{Atomic Visual Entailment: Enhancing Zero-Shot Vision--Language Reasoning through Atomic Fact Decomposition and Learned Selection}

\author{
  \textbf{Nallathambi Vethiappan\textsuperscript{1}},
  \textbf{Derya Soydaner\textsuperscript{1}},
  \textbf{Gijs Wijnholds\textsuperscript{1}}
\\
  \textsuperscript{1}Leiden Institute of Advanced Computer Science (LIACS), Leiden University, The Netherlands
\\
  \small{\textbf{Correspondence:} \texttt{nvethiappan@gmail.com}, \texttt{\{d.soydaner, g.j.wijnholds\}@liacs.leidenuniv.nl}}
}

\begin{document}
\maketitle
\begin{abstract}
Visual entailment (VE) asks whether an image supports, contradicts, or leaves undecided a textual hypothesis. Strong results come from fine-tuning large vision--language models on labelled data, while zero-shot and hybrid approaches remain far behind. A VE hypothesis often bundles several visual claims, yet existing zero-shot methods reason over it as a single unit. We propose \textbf{Atomic Visual Entailment} (AVE), which decomposes the hypothesis into atomic facts, produces candidate predictions from both the full hypothesis and its facts using frozen vision--language models, and predicts the final label with a lightweight classifier trained only on how those candidates behave. We find that decomposition helps only when the hypothesis context is preserved: judging facts in isolation is worse than not decomposing at all. Full-hypothesis and atomic prediction make complementary errors, and learning which to trust recovers far more of that complementarity than majority voting, reaching 0.803 test accuracy on SNLI-VE without fine-tuning any vision--language model. AVE also localises the visual evidence behind its prediction without region-level supervision. These results suggest that learning which candidate prediction to trust can close much of the gap to fine-tuned systems, offering a practical alternative where fine-tuning a vision-language model directly would need more labelled data or compute than is available.
\end{abstract}

\section{Introduction}
\label{sec:intro}

\begin{figure}[t]
\centering
\includegraphics[width=\linewidth]{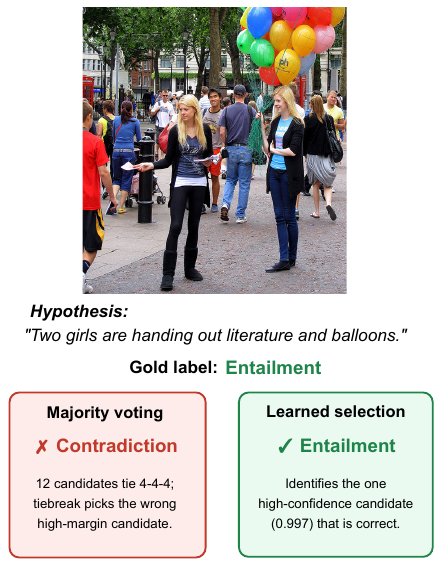}
\caption{AVE resolves prediction-pool disagreement that majority voting cannot. When candidates split evenly across labels, majority voting's tiebreak selects a confidently wrong candidate, while learned selection identifies the correct one.}
\label{fig:teaser}
\end{figure}

Vision--language models (VLMs) now perform well across many multimodal tasks, including judging whether an image supports a written claim. Visual entailment (VE) poses this as a three-way judgement over an image premise and a textual hypothesis: entailment, neutral, or contradiction \citep{xie2019visual}. It underpins multimodal fact-checking and misinformation detection, where an image may carry a wrong caption, be used out of context, or support only part of what is claimed \citep{akhtar-etal-2023-multimodal}. However, strong results on VE come from fine-tuning large VLMs on labelled data, and zero-shot and hybrid approaches remain comparatively unexplored.

Recent zero-shot and hybrid work on VE adds intermediate reasoning steps between the image and the label. IdealGPT \citep{you-etal-2023-idealgpt} generates sub-questions for a VLM to answer in a zero-shot setting, and Inner Monologue Multi-Modal Optimization (IMMO) \citep{yang2024tackling} trains a reasoner and an observer to alternate over several turns. In both, the intermediate steps interrogate the image rather than the claim. Yet a VE hypothesis often contains several visual claims, and reasoning over it as a single unit checks them all at once.

A natural response is to decompose the hypothesis into atomic facts, so that each claim can be checked against the image separately. Hypothesis decomposition has been used in Natural Language Inference (NLI), the textual counterpart of VE, to inspect model decisions at the level of individual claims, and that work finds the global context still matters: claims are often only interpretable when read together \citep{stacey-etal-2024-atomic, srikanth-rudinger-2025-nli}. Whether the same holds in VE, where each fact must be checked against an image rather than a textual premise, has not been studied.

A second challenge is deciding among competing predictions. A hypothesis can be judged as a whole or through its atomic facts, and instead of choosing one over the other, you can combine them to improve the prediction. Majority voting combines them without any training, but treats every candidate as equally reliable, so a tied or near-tied pool can be resolved in favour of a confidently wrong candidate (Figure~\ref{fig:teaser}). Related work on black-box ensembling instead learns which candidate to trust rather than counting votes, and does so without access to the models that produced the candidates \citep{xia-etal-2025-learn}.

To address these two challenges, we propose \textbf{Atomic Visual Entailment} (AVE), a framework that decomposes the hypothesis into atomic facts and combines predictions from both views of the claim. AVE produces a pool of candidate predictions from two frozen VLMs under several prediction strategies and prompt styles. A lightweight classifier then predicts the final label from features describing how those candidates behaved. Because each candidate carries the reasoning that produced it, AVE also localises the visual evidence behind its prediction, without region-level supervision. On SNLI-VE \citep{xie2019visual}, AVE outperforms every previously published zero-shot and hybrid result, while every VLM remains frozen.

In summary, our contributions are: (1) AVE, a framework that decomposes VE hypotheses into atomic facts and combines full-hypothesis and atomic predictions through a lightweight learned selector, without fine-tuning any VLM; (2) 0.803 test accuracy on SNLI-VE, 14.2 points above the strongest hybrid baseline and within the range of supervised systems; (3) Evidence that atomic reasoning helps only with hypothesis context preserved, and that learned selection exploits the resulting complementary errors far better than majority voting; (4) Zero-shot grounding of predicted evidence, reaching Recall@1 of 0.837 against human annotations without region-level supervision.

\section{Related Work}
\label{sec:related}

\textbf{Supervised visual entailment.} VE was introduced together with SNLI-VE and the EVE model, which fuses image and hypothesis features to predict the three-way relation \citep{xie2019visual}. Later work adapts large pretrained VLMs to the task, including UNITER \citep{chen2020uniter} and OFA \citep{wang2022ofa}, which reports the strongest published accuracy on SNLI-VE. These models require labelled SNLI-VE data and task-specific fine-tuning, which limits their use where such data is unavailable.

\textbf{Zero-shot and hybrid VE.} Zero-shot approaches avoid labelled VE data entirely. CLIP-based methods treat VE as image--text matching \citep{song-etal-2022-clip}, UniFine uses fine-grained visual and textual cues \citep{sun-etal-2023-unifine}, and IdealGPT decomposes the input into sub-questions answered by a VLM \citep{you-etal-2023-idealgpt}. Hybrid approaches combine zero-shot components with a small component trained on VE data: IMMO trains a reasoner and an observer to exchange questions and answers over several turns \citep{yang2024tackling}.

\textbf{Atomic decomposition.} Decomposing a sentence into atomic claims has been studied for textual NLI as a way to inspect model decisions at the level of individual claims. One direction decomposes the premise and derives an instance label from fact-level predictions \citep{stacey-etal-2024-atomic}, while another decomposes the hypothesis to study whether models predict consistently across a claim and its parts \citep{srikanth-rudinger-2025-nli}. Atomic inference is harder than sentence-level inference, and requiring every atomic fact to be entailed does not hold reliably in practice \citep{huang2026atomic}. All of this work concerns textual premises.

\textbf{Black-box prediction selection.} LLMs and VLMs are treated as black-box systems when their weights or confidence scores are unavailable, making fine-tuning difficult or costly \citep{xia-etal-2025-learn}. InfoSel addresses this by training a small selector to choose among candidate answers from several black-box models, using the question text, the candidate answers, and image region features \citep{xia-etal-2025-learn}. Its selector therefore encodes the original inputs alongside the candidates, and its output is always one of them. AVE uses the same idea for VE, but its selector is trained only on how the candidate pool behaves and predicts the label directly.

\textbf{Evidence grounding.} Fine-grained VE localises the visual evidence behind a prediction, but relies on region-level annotation collected for that purpose \citep{thomas2022finegrained}. Open-vocabulary detectors localise arbitrary text queries without task-specific training \citep{liu2024grounding}, which removes the need for such annotation.

\section{Method}
\label{sec:method}

\begin{figure*}[t]
\centering
\includegraphics[width=1\textwidth]{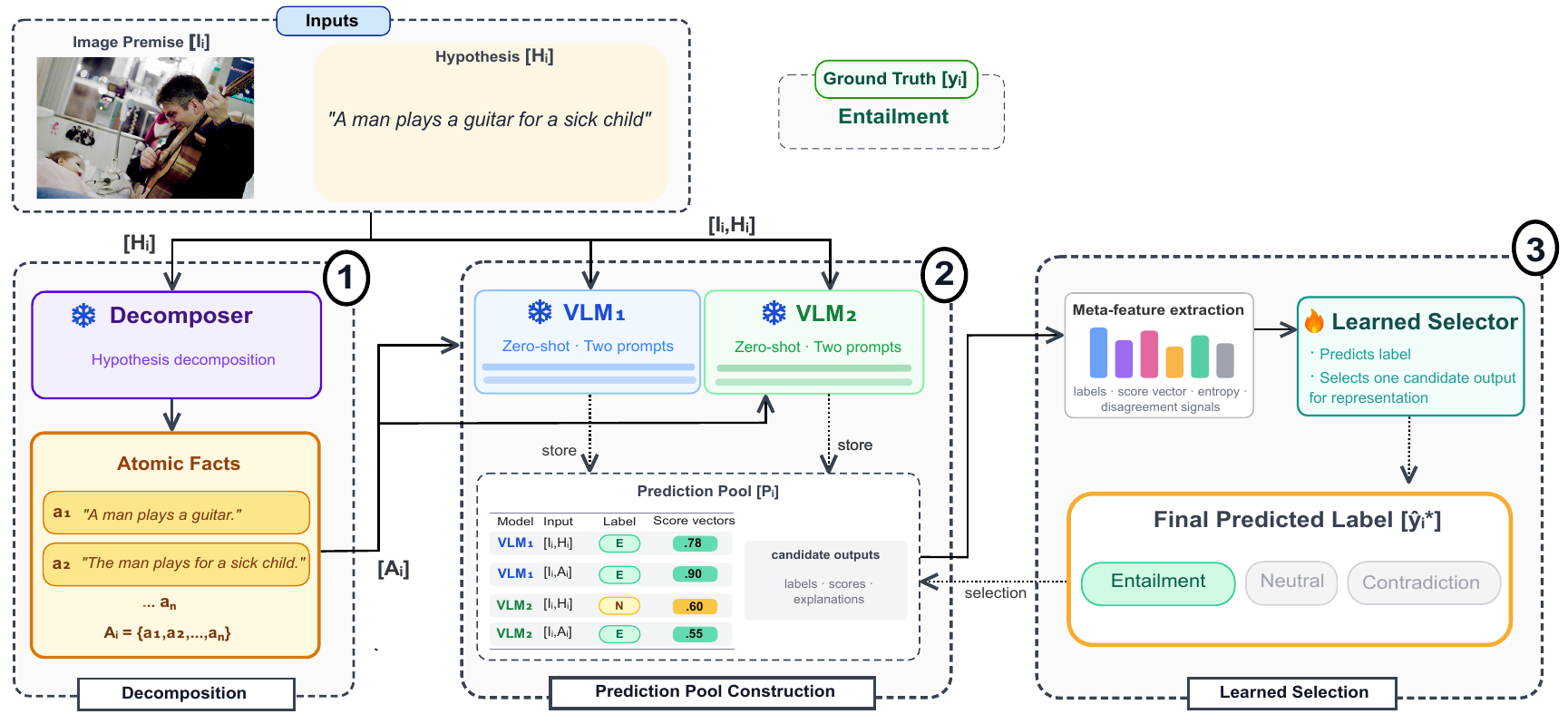}
\caption{The AVE framework. Components marked with a snowflake are frozen; the learned selector is the only trained component. Given an image premise and a
hypothesis, the \texttt{Decomposer} generates atomic facts, frozen VLMs produce candidate label--score pairs from the full hypothesis and from the atomic facts, the \texttt{Selector} predicts the final label and selects a representative candidate for grounding.}
\label{fig:pipeline}
\end{figure*}

In this section, we introduce the proposed \textbf{AVE}. Our focus is the task of VE, where a hypothesis $H$ is judged against an image $I$ with a label from $\mathcal{Y} = \{\textit{entailment}, \textit{neutral}, \textit{contradiction}\}$. AVE consists of four components: a \texttt{Decomposer}, a \texttt{Prediction Pool}, a \texttt{Selector}, and a \texttt{Grounder}. Given $H$, the \texttt{Decomposer} splits the hypothesis into self-contained atomic facts $A = \{a_1, \dots, a_n\}$, so that a claim bundling several assertions can be checked at a finer granularity (Sec.~\ref{sec:decomposition}). The \texttt{Prediction Pool} applies several prediction strategies over $H$, $A$, and $I$ using frozen VLMs, producing candidate predictions $P = \{c_1, \dots, c_K\}$ (Sec.~\ref{sec:pool}). The objective is not to identify the single best strategy, but to obtain candidates that disagree in informative ways: at least one candidate is correct far more often than any individual strategy, which motivates treating the final decision as a selection problem. The \texttt{Selector} therefore predicts the final label $\hat{y}$ by casting this decision as classification over lightweight meta-features describing the candidates, their label likelihoods, and their agreement (Sec.~\ref{sec:selector}). Finally, the \texttt{Grounder} localises phrases from the VLM's reasoning behind the selected candidate, tying the decision to visible evidence (Sec.~\ref{sec:grounding}). No VLM is fine-tuned at any stage. The overall pipeline is shown in Figure~\ref{fig:pipeline}.

\subsection{Hypothesis Decomposition}
\label{sec:decomposition}

The \texttt{Decomposer} converts a hypothesis $H$ into an atomic fact set $A = \{a_1, \dots, a_n\}$, where $n$ varies with the number of distinct claims in $H$. If $H$ already expresses a single claim it is kept as one atomic fact; otherwise it is split into the fewest facts needed to cover its meaning. Each fact is self-contained, with unclear pronouns replaced by explicit noun phrases, and uses only information stated in $H$ (Figure~\ref{fig:decompose}).

We use Qwen2.5-32B-Instruct\footnote{\url{https://huggingface.co/Qwen/Qwen2.5-32B-Instruct}} as the \texttt{Decomposer}, selected from a comparison of text-only LLMs validated manually by an author on 100 sampled hypotheses (Appendix \ref{app:decomposer}). For each hypothesis, BM25 \citep{trotman2014improvements} ranks a manually constructed pool of 50 hypothesis--atomic fact pairs by lexical similarity to $H$, and the four highest-ranked pairs are supplied as demonstrations. Decomposition is text-only and independent of the image, so $A$ is computed once per hypothesis and reused by every strategy in the prediction pool.

\subsection{Candidate Prediction Pool}
\label{sec:pool}

Judging $H$ directly preserves the relations between its claims, while judging the atomic facts $A$ exposes each claim separately. The \texttt{Prediction Pool} retains both views rather than committing to one, producing a set of $K$ candidate predictions $P = \{c_1, \dots, c_K\}$.

We use two frozen VLMs, Qwen3-VL-8B\footnote{\url{https://huggingface.co/Qwen/Qwen3-VL-8B-Instruct-hf}} and InternVL3-8B\footnote{\url{https://huggingface.co/OpenGVLab/InternVL3-8B}}, selected from a comparison of seven models on the accuracy--size trade-off (Appendix~\ref{app:vlm}). Each VLM sees the image $I$ in every case and applies three zero-shot prediction strategies, differing in how the hypothesis is presented alongside it. \textit{Full-hypothesis} prediction presents $H$ as a single statement and asks for one label (Figure~\ref{fig:fullhyp}). \textit{Atomic} prediction replaces $H$ with the facts in $A$, presented together as parts of one hypothesis rather than as independent claims, so the model reasons over the individual claims without losing the claim they compose (Figure~\ref{fig:atomic}). \textit{Self-decomposed atomic} prediction presents $H$ alone. The model decomposes the hypothesis itself and labels the resulting facts within a single generation, so unlike atomic prediction, it does not use $A$ (Figure~\ref{fig:selfdecomp}). A fourth strategy, independent atomic, judges each fact in isolation (Figure~\ref{fig:ind}) and combines the per-fact labels by rule. $H$ is \texttt{entailment} when every $a \in A$ is \texttt{entailment}, \texttt{contradiction} when at least one $a \in A$ is \texttt{contradiction}, and \texttt{neutral} otherwise. We exclude it from the pool and return to why in Section~\ref{sec:analysis}.

Each strategy is run under two prompt styles, motivated by the sensitivity of zero-shot models to prompting \citep{pitta-etal-2025-probing}. The \textit{simple} prompt asks for a label and a short justification for that label. The \textit{structured} prompt asks the model to identify the relevant visual evidence and reason about whether it supports, contradicts, or is insufficient for the statement before returning a label. For the atomic strategies, it also asks for a per-fact judgement along with a
single final label. The prompt states how to combine those judgements: contradiction if any fact conflicts with the image, entailment if all are
supported, and neutral otherwise. Two VLMs, three strategies, and two prompt styles give $K = 12$ candidates per instance.

From each candidate, we extract two labels, a label distribution, and a justification. The VLM's response contains the \textit{generated label} $\hat{y}^{g}_k$, the label the model states directly, together with a justification $e_k$. Separately, and from the same prompt $x_k$, we score each label $y$ as a forced continuation \citep{brown2020language}, masking the prompt tokens so that only the label contributes:

\begin{equation}
s_k(y) = \sum_{t=1}^{|y|} \log p_\theta(y_t \mid I, x_k, y_{<t})
\label{eq:score}
\end{equation}

where $y \in \mathcal{Y}$ and $y_t$ are its tokens. A softmax over $\{s_k(y)\}$ gives a distribution $\tilde{s}_k$, whose argmax is the \textit{score-based label} $\hat{y}^{s}_k$. This reflects the model's relative preference across the label set rather than the label it verbalised, and the two need not agree.

\subsection{Combining Candidate Predictions}
\label{sec:selector}

\textbf{Majority voting.} A training-free way to derive the final label from the prediction pool is majority voting (AVE-MV), which takes the most frequent generated label across its candidates, breaking ties by the largest score margin, the gap between a candidate's top two label-likelihood scores from Equation~\ref{eq:score}. Each candidate's vote counts the same regardless of how accurate that candidate is overall.

\textbf{Learned selection.} At least one candidate in the prediction pool already matches the gold label for over 94\% of instances (Table~\ref{tab:main}), so the remaining problem is identifying which candidate to trust rather than re-examining the image or hypothesis. AVE-LS therefore casts the final decision as classification over meta-features describing how the pool behaved on a given input, rather than a model that also processes the image or text. For instance $i$, we form a feature vector $x_i$ whose features fall into four groups. \textit{Per-candidate} features record each $c_k$'s generated and score-based labels and its score vector $\tilde{s}_k$. From $\tilde{s}_k$ we also derive the highest score, its margin over the second, and its entropy. \textit{Pool-level} features summarise the label distribution across candidates, such as how often each label is predicted and whether more than half the candidates agree on one. \textit{Pairwise agreement} features record whether each pair of candidates agrees, separately for generated and score-based labels. \textit{Decomposition metadata} records the number of atomic facts and the lengths of $H$ and $A$. The feature vector contains no image, no hypothesis or atom text, and none of the candidates' justifications, making AVE-LS a meta-classifier over prediction-pool behaviour rather than over the original inputs.

Tree-based classifiers handle this kind of mixed-scale tabular input directly, without the feature scaling or larger training sets that a neural classifier would need. We compare two tree-based classifiers, HistGradientBoosting and XGBoost, and select XGBoost for its higher validation macro-F1; its hyperparameter configuration is given in Appendix~\ref{app:selector}. We train this lightweight classifier $f_\phi$ \citep{chen2016xgboost} on the training split to predict the gold label from $x_i$, giving $\hat{y} = f_\phi(x_i)$ with $\hat{y} \in \mathcal{Y}$. Because the target is the label rather than the identity of a candidate, $\hat{y}$ is not restricted to the labels present in $P$. We compare the two objectives in Section~\ref{sec:analysis}.

\textbf{Representative candidate.} Predicting a label directly does not identify which candidate in $P$ supports it. When at least one candidate has
generated label $\hat{y}^{g}_k = \hat{y}$, we select the one whose score for $\hat{y}$ is highest and take its justification $e_k$ as the representative
output for that instance. Otherwise, the label is returned without a representative justification.

\subsection{Evidence Grounding}
\label{sec:grounding}

AVE's prediction includes a textual justification that describes the visual evidence but does not indicate where it appears in the image. The \texttt{Grounder} localises that evidence in two steps: it extracts a phrase from the justification $e_k$ of the representative candidate, then locates that phrase in the image. Only entailment and contradiction predictions are grounded, since their justifications point to visible content that settles the claim. A neutral justification explains why the image leaves the claim undecided, so no region can be localised.

Phrase extraction takes $H$, $\hat{y}$, and $e_k$, and returns a short phrase naming visible content. For entailment, the phrase names content that supports the claim. For contradiction, it names what the image shows instead, since the contradicting evidence is what is visible rather than what $H$ asserts. We use Qwen3-8B as the extractor (Figure~\ref{fig:phrase}).

Grounding DINO \citep{liu2024grounding}, using the grounding-dino-base checkpoint, then takes the image $I$ and the phrase as a text query and returns bounding boxes for the matching regions. The boxes accompany the predicted label as its visual evidence, and $\hat{y}$ itself is unchanged.

\section{Experiments}
\label{sec:experiments}

We evaluate AVE on SNLI-VE. We first describe the setup, then report what the decomposition produces and how the individual prediction strategies perform. We then compare majority voting with learned selection, both against the prediction pool itself and against prior work. Finally, we evaluate the grounded evidence against human annotations.

\subsection{Experimental Setup}
\label{sec:setup}

\textbf{Datasets.} We evaluate on SNLI-VE \citep{xie2019visual}, which pairs Flickr30k images with SNLI hypotheses labelled entailment, neutral, or contradiction, and follows the official splits of 17,858 development and 17,901 test instances. Candidate predictions are also generated for a 47,203-instance subset of the training split, which is used to train the \texttt{Selector}. Grounding is evaluated against Flickr30k Entities \citep{flickrentitiesijcv}, which provides human-annotated bounding boxes linking image regions to noun phrases in the corresponding captions. We sample 2,500 test instances per groundable label for grounding evaluation.

\textbf{Models.} All models are used with greedy decoding, so repeated runs give identical output. Generation is capped at 700 tokens, which accommodates the longest structured responses. Grounding DINO uses a box threshold of 0.35 and a text threshold of 0.25, and we retain the three highest-scoring boxes after suppressing near-duplicates.

\textbf{Selector training.} AVE-LS is trained on the candidate predictions for the training subset, holding out 20\% for model selection. The classifier and its hyperparameters are chosen by validation macro-F1 (Appendix~\ref{app:selector}). The development and test splits are used once, after this configuration is fixed.

\textbf{Metrics.} We report accuracy and macro-F1 for label prediction. We additionally report an oracle upper bound: the proportion of instances for which at least one candidate in the pool matches the gold label, which is the accuracy an ideal selector over that pool would reach. For grounding, we report Recall@1 and Recall@3 at an intersection-over-union (IoU) threshold of 0.5, together with the mean best IoU among the three retained boxes. Since Flickr30k Entities annotates caption phrases, not hypothesis evidence, these scores measure localisation quality but only approximate correctness.

\textbf{Compute.} All experiments run on a university HPC cluster. The 32B Decomposer runs on an NVIDIA A100 GPU with 80GB memory. All 8B VLMs run on an NVIDIA A100 MIG instance with 40GB memory. The Selector trains on CPU, completing in under two minutes per configuration.

\subsection{Decomposition}
\label{sec:decomp-results}

Table~\ref{tab:atoms} reports the atom-count distribution on the test split, which is stable across splits. Most hypotheses decompose into one or two atomic facts. The label composition shifts as the atom count grows. Neutral rises from 27.4\% of single-atom hypotheses to over 49\% with three or more, while entailment falls from 39.2\% to 20.4\%. Hypotheses labelled neutral therefore tend to contain several separate claims, which decomposition breaks apart.

\begin{table}[h]
\centering
\begin{tabular}{lrrrr}
\toprule
\textbf{Atoms} & \textbf{Share} & \textbf{E} & \textbf{N} & \textbf{C} \\
\midrule
1  & 63.6 & 39.2 & 27.4 & 33.4 \\
2  & 31.8 & 23.6 & 42.6 & 33.9 \\
3  & 3.9  & 20.4 & 51.1 & 28.5 \\
4+ & 0.7  & 25.0 & 49.2 & 25.8 \\
\bottomrule
\end{tabular}
\caption{Atomic fact distribution on the test split. \textit{Share} is the
percentage of instances in each bucket; E, N and C are gold-label percentages
within it.}
\label{tab:atoms}
\end{table}

\begin{figure*}[t]
\centering
\begin{subfigure}{0.53\textwidth}
  \includegraphics[width=0.95\linewidth]{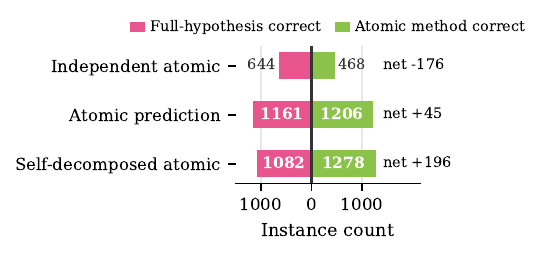}
  \caption{}
\end{subfigure}
\hfill
\begin{subfigure}{0.46\textwidth}
  \includegraphics[width=0.95\linewidth]{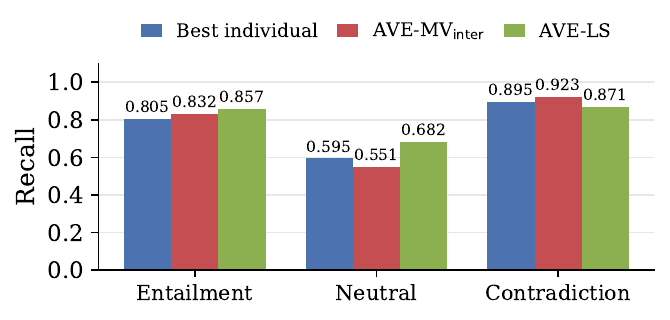}
  \caption{}
\end{subfigure}
\caption{(a) Instance-level disagreement with full-hypothesis prediction on test. Independent atomic prediction is excluded from the prediction pool. (b) Per-class recall on test.}
\label{fig:behaviour}
\end{figure*}

\begin{table}[t]
\centering
\small
\begin{tabular}{@{}l@{\hskip 0.35em}cccc@{}}
\toprule
& \multicolumn{2}{c}{Development} & \multicolumn{2}{c}{Test} \\
\cmidrule(lr){2-3} \cmidrule(lr){4-5}
Method & Acc. & F1 & Acc. & F1 \\
\midrule
Best full-hypothesis        & 0.749 & 0.732 & 0.754 & 0.737 \\
Best atomic           & 0.750 & 0.739 & 0.757 & 0.746 \\
Best self-decomposed        & 0.762 & 0.757 & 0.765 & 0.760 \\
\midrule
AVE-MV$_\text{intra}$ (Qwen3)     & 0.752 & 0.744 & 0.756 & 0.748 \\
AVE-MV$_\text{intra}$ (InternVL3) & 0.761 & 0.751 & 0.768 & 0.757 \\
AVE-MV$_\text{inter}$             & 0.765 & 0.756 & 0.769 & 0.760 \\
\textbf{AVE-LS}             & \textbf{0.799} & \textbf{0.798} & \textbf{0.803} & \textbf{0.801} \\
\midrule
Oracle ($K=12$)             & 0.941 & 0.941 & 0.941 & 0.940 \\
Oracle (chance pool)             & 0.992 & 0.992 & 0.992 & 0.992 \\
\bottomrule
\end{tabular}
\caption{Best individual candidate per prediction strategy, majority voting, and learned selection. AVE-MV\textsubscript{intra} votes among the six candidates from one VLM; AVE-MV\textsubscript{inter} votes across all twelve candidates from both VLMs. Oracle reports the proportion of instances for which at least one of the $K=12$ candidates matches the gold label; the chance pool replaces those candidates with twelve independent random labels.}
\label{tab:main}
\end{table}

\subsection{Prediction Pool and Combination}
\label{sec:main-results}

Table~\ref{tab:main} reports the best individual candidate for each prediction strategy, together with the two combination methods and the oracle upper bound.

\textbf{Individual strategies.} The three strategies span 1.3 points of development accuracy, from 0.749 for full-hypothesis prediction to 0.762 for self-decomposed atomic prediction. Both atomic strategies exceed full-hypothesis prediction, and self-decomposition is strongest on both splits,
but the margins are too small for any strategy to be clearly preferable on accuracy alone.

\textbf{Complementary errors.} Atomic prediction differs from full-hypothesis prediction by 0.3 accuracy points on test, yet the two disagree
on 2,367 instances (Figure~\ref{fig:behaviour}a). Self-decomposed prediction disagrees on a comparable number but recovers substantially more than it loses.
The strategies therefore fail on largely different instances. At least one candidate matches the gold label for 94.1\% of test instances, bounding what
any selector over this pool can achieve. High coverage is not by itself informative: twelve random labels would cover the gold label 99.2\% of the time, yet selecting among them would be no better than chance. The pool is useful because its candidates also carry signal about which of them to trust.

\textbf{Majority voting.} AVE-MV$_\text{inter}$ reaches 0.765 on development and 0.769 on test, improving over the best individual candidate by 0.3 and 0.4 points. Voting within a single VLM is weaker still, with AVE-MV$_\text{intra}$ (Qwen3) falling below the best individual candidate on both splits. Treating every candidate as equally reliable therefore recovers almost none of the gap to the oracle.

\textbf{Learned selection.} AVE-LS reaches 0.799 on development and 0.803 on test, improving over AVE-MV$_\text{inter}$ by 3.4 points on both splits. On test, voting gains 0.4 points over the best individual candidate, while learned selection gains 3.8. A lightweight classifier over prediction-pool behaviour is therefore enough to exploit the signal that voting discards. The gain is concentrated on neutral, whose test recall rises from 0.551 to 0.682 at a 5.2-point cost to contradiction (Figure~\ref{fig:behaviour}b), the label that improves the most yet remains the hardest under every method. The full confusion matrices behind this trade-off are given in Appendix~\ref{app:confusion}.

\subsection{Comparison with Prior Work}
\label{sec:prior}

\begin{table}[t]
\centering
\begin{tabular}{@{}llcc@{}}
\toprule
& \textbf{Method} & \textbf{Dev} & \textbf{Test} \\
\midrule
\multirow{3}{*}{\rotatebox{90}{\small Sup.}}
& \textcolor{gray}{EVE-Image \citep{xie2019visual}}    & \textcolor{gray}{0.716} & \textcolor{gray}{0.711} \\
& \textcolor{gray}{UNITER \citep{chen2020uniter}} & \textcolor{gray}{--}   & \textcolor{gray}{0.794} \\
& \textcolor{gray}{OFA \citep{wang2022ofa}}         & \textcolor{gray}{0.910} & \textcolor{gray}{0.912} \\
\midrule
\multirow{3}{*}{\rotatebox{90}{\small ZS}}
& CLIP \citep{song-etal-2022-clip}   & 0.676 & 0.682 \\
& IdealGPT \citep{you-etal-2023-idealgpt} & 0.553 & --   \\
& AVE-MV$_\text{inter}$ (ours)   & 0.765 & 0.769 \\
\midrule
\multirow{2}{*}{\rotatebox{90}{\small Hyb.}}
& IMMO \citep{yang2024tackling} & 0.657 & -- \\
& \textbf{AVE-LS} \textbf{(ours)}     & \textbf{0.799} & \textbf{0.803} \\
\bottomrule
\end{tabular}
\caption{Accuracy on SNLI-VE (Sup: Supervised, ZS: Zero-Shot, Hyb: Hybrid).}
\label{tab:prior}
\end{table}

Table~\ref{tab:prior} reports accuracy on SNLI-VE against previously published results. Among zero-shot methods, AVE-MV$_\text{inter}$ is the strongest on SNLI-VE, improving over IdealGPT by 21.2 points on development (76.5\% vs 55.3\%) without the proprietary reasoning backbone IdealGPT relies on. AVE-LS is the strongest result among zero-shot and hybrid methods, improving over IMMO by 14.2 points (79.9\% vs 65.7\%). Both use labelled VE data, but IMMO fine-tunes a 7B model, updating roughly five million parameters through LoRA. In contrast, AVE-LS leaves both 8B VLMs frozen and trains a gradient-boosted classifier over their outputs.

Supervised results are reported as a reference. AVE-LS exceeds EVE-Image and matches UNITER, though a substantial gap to OFA remains. Fine-tuning a large end-to-end VLM on the full SNLI-VE training set is clearly effective when labelled data and compute are available. AVE-LS reaches comparable accuracy to mid-range supervised systems without updating any VLM weights, which matters when fine-tuning is impractical or when the reasoning behind a prediction needs to remain inspectable.

\subsection{Evidence Grounding}
\label{sec:grounding-results}

\begin{table}[t]
\centering
\small
\begin{tabular}{@{}lrrr@{}}
\toprule
& \textbf{Entail.} & \textbf{Contra.} & \textbf{Overall} \\
\midrule
Matched      & 2,173 & 1,933 & 4,106 \\
Coverage     & 0.869  & 0.773  & 0.821 \\
Recall@1     & 0.823  & 0.852  & 0.837 \\
Recall@3     & 0.896  & 0.904  & 0.900 \\
Mean IoU@3   & 0.843  & 0.847  & 0.845 \\
\bottomrule
\end{tabular}
\caption{Grounding against Flickr30k Entities. Recall uses IoU $\geq 0.5$.}
\label{tab:grounding}
\end{table}

Table \ref{tab:grounding} reports grounding results on the sampled test instances. An extracted phrase matches a Flickr30k Entities annotation for 82.1\% of them; the remainder describe evidence that the caption-based annotations do not name. Coverage is lower for contradiction than for entailment, at 77.3\% against 86.9\%, since contradiction evidence often concerns an absent object or a mismatched attribute rather than a named entity.

\begin{figure*}[t]
\centering
\includegraphics[width=0.75\textwidth]{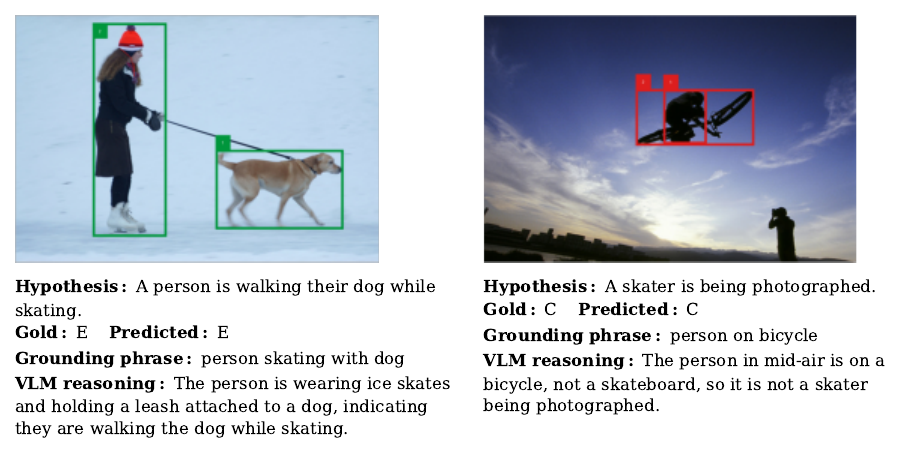}
\caption{Grounding examples. For entailment, the phrase names visible evidence supporting the hypothesis; for contradiction, it names what the image shows instead. E and C denote entailment and contradiction.}
\label{fig:grounding-examples}
\end{figure*}

Among matched instances, Recall@1 reaches 0.837, and Recall@3 reaches 0.900, with a mean best IoU of 0.845 among the three retained boxes. Fine-grained VE has previously relied on region-level annotations to produce visual evidence \citep{thomas2022finegrained}. AVE reaches comparable localisation quality from reasoning the VLM has already produced, with no component trained for the task, as the entailment and contradiction examples in Figure~\ref{fig:grounding-examples} illustrate.

\subsection{Ablation Studies}
\label{sec:analysis}

\textbf{Independent atomic prediction.} Judging each atomic fact in isolation and combining the per-fact labels by rule is the only strategy that performs worse than full-hypothesis prediction. It recovers 468 test instances where full-hypothesis prediction fails, but loses 644 where it succeeds (Figure~\ref{fig:behaviour}a). Isolating a fact removes its context, and an error on any fact propagates to the final label with no recovery. The same pattern holds in textual NLI, where entailment of every atomic fact does not reliably yield the sentence-level label \citep{huang2026atomic}. Atomic reasoning needs shared context among the facts, so AVE presents them together. 

\textbf{Prediction pool composition.} We compare the full $K=12$ pool against three single-family variants, each with four candidates from one strategy or prompt style. The full pool is best at every training size (Appendix~\ref{app:efficiency}). Trained on 5,000 examples, it reaches 0.787 validation macro-F1, above the full-hypothesis-only variant trained on all 37,762 (0.783), and gains the most from additional data, while single-family variants plateau earlier and lower. The advantage therefore comes from combining strategies rather than from any single one.

\textbf{Label-level versus candidate-level selection.} We compare predicting the label directly against selecting a candidate and inheriting its label. Using the same features, split and classifier, the label-level objective reaches 0.801 validation accuracy against 0.778 for the candidate-level objective. Choosing among twelve candidates that differ in model, prompt, and score is harder than choosing among three labels, so AVE-LS predicts the label directly.

\textbf{Prompt style sensitivity.} Neither prompt style dominates, and the effect depends on both the model and the prediction strategy. For Qwen3-VL, the structured prompt helps atomic prediction (0.716 to 0.745) and self-decomposed prediction (0.709 to 0.748). For InternVL3, it hurts full-hypothesis prediction (0.752 to 0.723), and self-decomposed prediction with score-based labels collapses to 0.365 under the structured prompt while reaching 0.613 under the simple one. Since no style is reliably better, AVE keeps both in the pool and leaves the selector to weigh them.

\textbf{Hypothesis complexity.} Accuracy declines for every strategy as the number of atomic facts grows, from 0.774 at one fact to 0.637 at four or more for full-hypothesis prediction (Figure~\ref{fig:complexity}). Part of this reflects label distribution rather than the strategies, since neutral grows more frequent at higher atom counts (Table~\ref{tab:atoms}) and remains hardest throughout. Independent atomic prediction falls behind from two facts onward, as isolated evaluation grows less reliable with more facts. Atomic and self-decomposed stay ahead of full-hypothesis up to three facts, after which atomic falls behind while self-decomposed remains strongest, though the four-or-more bucket holds under 1\% of instances.

\section{Conclusion}
\label{sec:conclusion}

We study whether atomic hypothesis decomposition improves zero-shot visual entailment, and whether predictions from different reasoning routes can be combined. We propose AVE, which decomposes the hypothesis into atomic facts, produces candidate predictions from both the full hypothesis and its facts using frozen VLMs, and predicts the final label with a lightweight classifier trained only on how those candidates behave. We find that decomposition helps only when the hypothesis context is preserved: judging facts in isolation performs worse than not decomposing at all. Full-hypothesis and atomic reasoning make complementary errors, and learning which to trust recovers substantially more of that complementarity than majority voting, reaching 0.803 test accuracy on SNLI-VE without fine-tuning any VLM. Because the selected prediction carries the reasoning that produced it, AVE also localises the visual evidence behind that prediction, without region-level supervision.

\section*{Limitations}

\textbf{Inference cost.} AVE evaluates twelve candidates per instance, so VLM calls scale linearly with pool size, and adding models, prompt styles, or strategies increases this further. The selector itself adds negligible overhead. Reducing the pool while retaining most of the benefit could reduce the inference cost. 

\textbf{Single dataset and language.} All experiments use SNLI-VE, an English-language VE benchmark built on Flickr30k captions. Our conclusions therefore apply to this evaluation setting. Validation across other entailment benchmarks, image domains, and languages is an important direction for future work.

\textbf{Approximate grounding evaluation.} SNLI-VE has no annotations linking hypothesis-level evidence to image regions, so we evaluate against Flickr30k Entities, whose annotations describe caption phrases rather than entailment evidence. Reported scores therefore approximate rather than directly measure grounding accuracy. A purpose-built annotation set is left to future work.

\textbf{Decomposition quality.} Decomposition quality was assessed manually by an author on 100 hypotheses, giving an initial estimate rather than a validated measure of faithfulness and completeness. A multi-annotator study could provide further validation. 

\textbf{Selector transparency.} AVE-LS's per-instance decisions are not directly interpretable; its accuracy was still rising at the largest training size tested, and it is trained on the specific behaviour of the two VLMs and prompts used to build the pool, so it may not transfer if either changes. These aspects need further investigation.

\section*{Ethics Statement}

SNLI-VE and Flickr30k Entities are publicly available research benchmarks and involve no new data collection from human subjects. Qwen3-VL-8B and InternVL3-8B are pretrained on large web corpora and may carry biases from that data into their predictions. Visual entailment is relevant to fact-checking and misinformation detection, where misplaced trust in model outputs could have real-world consequences, so predictions should be used as a decision aid reviewed by a person rather than as an automated final judgement of whether a claim is true.


\bibliography{custom}

\newpage

\appendix

\section{Decomposer Comparison}
\label{app:decomposer}

Table~\ref{tab:decomposer} compares three text-only LLMs as the Decomposer. All three are faithful, and the difference lies in completeness. Qwen2.5-32B-Instruct scores highest on completeness and is used throughout.

\begin{table}[H]
\centering
\small
\begin{tabular}{@{}lrrr@{}}
\toprule
Decomposer & Faith. & Compl. & Both \\
\midrule
Meta-Llama-3.1-8B & 95 & 88 & 88 \\
Qwen3-8B          & 96 & 90 & 90 \\
Qwen2.5-32B       & \textbf{98} & \textbf{96} & \textbf{96} \\
\bottomrule
\end{tabular}
\caption{Decomposition quality on 100 hypotheses sampled from the training split, annotated manually by an author. Faithfulness checks that no fact adds information absent from the hypothesis; completeness checks that all claims are covered. Counts out of 100.}
\label{tab:decomposer}
\end{table}

\section{Vision--Language Model Comparison}
\label{app:vlm}

\begin{figure}[H]
\centering
\includegraphics[width=\linewidth]{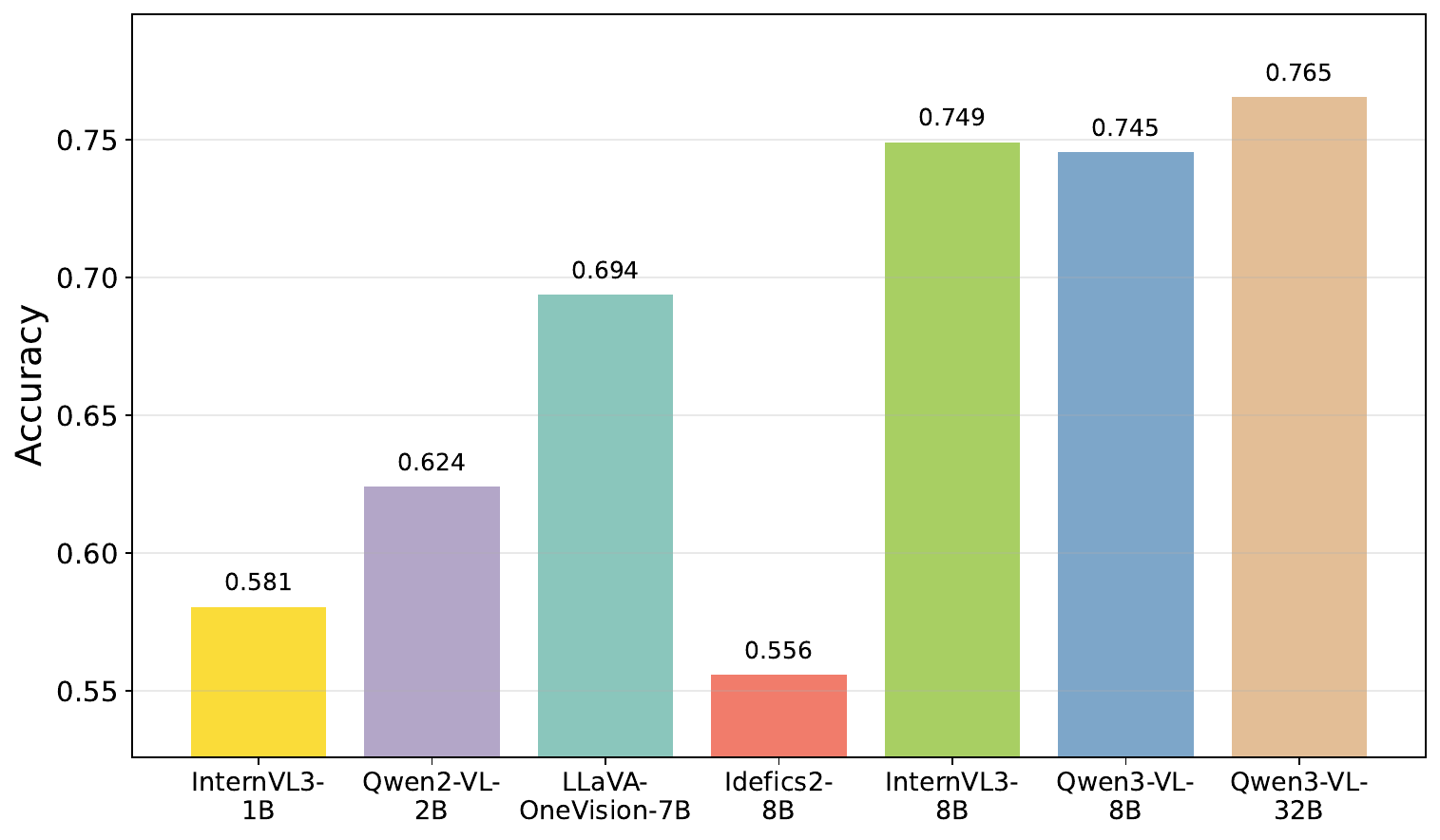}
\caption{Full-hypothesis accuracy on the development split for seven VLMs, using each model's best prompt style and label derivation.}
\label{fig:vlm}
\end{figure}

Figure~\ref{fig:vlm} compares seven VLMs on full-hypothesis prediction. Qwen3-VL-32B reaches the highest accuracy at 0.765, ahead of InternVL3-8B (0.749) and Qwen3-VL-8B (0.745). The gain over the two 8B models is only 1.6 to 2.0 points despite roughly four times the parameters. Idefics2-8B performs worst overall, below even the 1B and 2B models. AVE retains InternVL3-8B and Qwen3-VL-8B for their accuracy-size balance.

\section{Selector Configuration}
\label{app:selector}

Table~\ref{tab:variants} lists the four feature variants compared during selector training. Feature count grows with $K$, since per-candidate and pairwise agreement features scale with the number of candidates.

\begin{table}[H]
\centering
\small
\begin{tabular}{@{}lrr@{}}
\toprule
Feature variant & $K$ & Features \\
\midrule
Full-hypothesis    & 4  & 129 \\
Simple atomic      & 4  & 129 \\
Structured atomic  & 4  & 129 \\
Full 12 methods    & 12 & 409 \\
\bottomrule
\end{tabular}
\caption{Feature variants compared during AVE-LS training. $K$ is the number of candidates included.}
\label{tab:variants}
\end{table}

\noindent AVE-LS uses XGBoost, selected against HistGradientBoosting across four feature variants by validation macro-F1 (Table~\ref{tab:selector_comparison}). Training uses a fixed stratified 80/20 split of the training subset with seed 42.

\begin{table}[H]
\centering
\small
\begin{tabular}{@{}lrrrr@{}}
\toprule
& \multicolumn{2}{c}{HistGB} & \multicolumn{2}{c}{XGBoost} \\
Feature variant & Acc. & F1 & Acc. & F1 \\
\midrule
Full-hypothesis    & 0.790 & 0.787 & 0.788 & 0.785 \\
Simple atomic      & 0.792 & 0.788 & 0.794 & 0.792 \\
Structured atomic  & 0.793 & 0.790 & 0.793 & 0.791 \\
Full 12 methods    & 0.798 & 0.796 & \textbf{0.801} & \textbf{0.798} \\
\bottomrule
\end{tabular}
\caption{Validation accuracy and macro-F1 for AVE-LS across classifier families and feature variants.}
\label{tab:selector_comparison}
\end{table}

\noindent Table~\ref{tab:hparams} reports the resulting hyperparameters for the selected XGBoost configuration.

\begin{table}[H]
\centering
\small
\begin{tabular}{@{}lr@{}}
\toprule
Parameter & Value \\
\midrule
objective          & multi:softprob \\
eval\_metric       & mlogloss \\
n\_estimators      & 1{,}100 \\
max\_depth         & 3 \\
learning\_rate     & 0.025 \\
subsample          & 0.88 \\
colsample\_bytree  & 0.88 \\
reg\_lambda        & 3.5 \\
reg\_alpha         & 0.25 \\
min\_child\_weight & 1.5 \\
gamma              & 0.02 \\
tree\_method       & hist \\
random\_state      & 42 \\
\bottomrule
\end{tabular}
\caption{XGBoost hyperparameters for AVE-LS, chosen by validation macro-F1.}
\label{tab:hparams}
\end{table}

\section{Data Efficiency}
\label{app:efficiency}

\begin{figure}[H]
\centering
\includegraphics[width=0.9\linewidth]{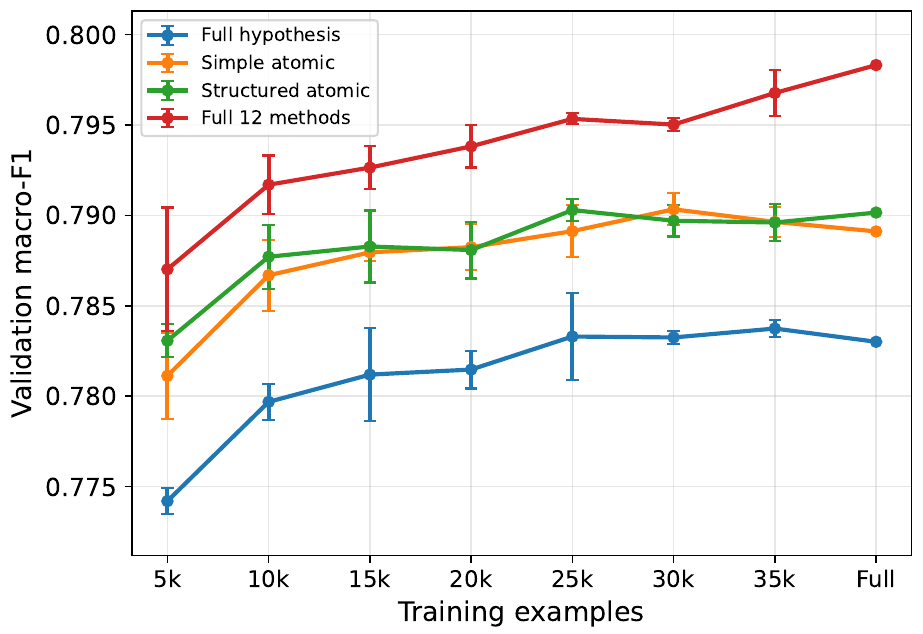}
\caption{Validation macro-F1 against selector training size for the four feature variants. Error bars show one standard deviation across three random training subsamples at each size.}
\label{fig:efficiency}
\end{figure}

Figure~\ref{fig:efficiency} shows validation macro-F1 against training size for the four feature variants, averaged over three random training subsamples at each size. The full pool leads at every size and gains the most from additional data, while the single-family variants plateau sooner and at a lower macro-F1. At 5,000 examples, the full pool already exceeds the full-hypothesis-only variant trained on all 37,762. The full pool continues to improve across the entire range without flattening, unlike the single-family variants, which level off well before the largest training sizes.

\section{Confusion Matrices}
\label{app:confusion}

\begin{figure}[H]
\centering
\includegraphics[width=0.9\linewidth]{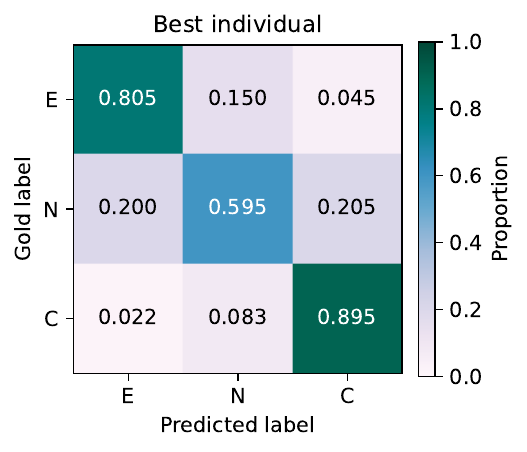}
\caption{Row-normalised confusion matrix for the best individual candidate on the test set.}
\label{fig:cm_best}
\end{figure}

\begin{figure}[H]
\centering
\includegraphics[width=0.9\linewidth]{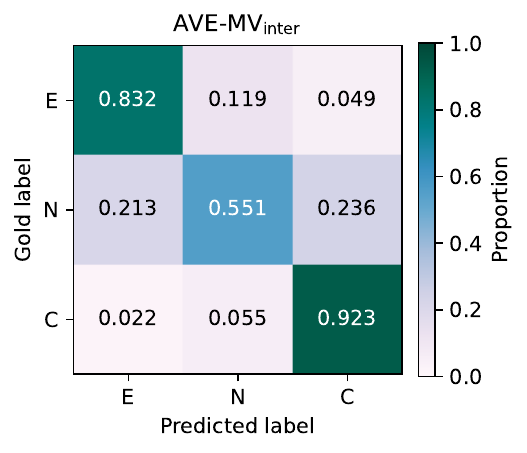}
\caption{Row-normalised confusion matrix for AVE-MV\textsubscript{inter} on the test set.}
\label{fig:cm_mv}
\end{figure}

\begin{figure}[H]
\centering
\includegraphics[width=0.9\linewidth]{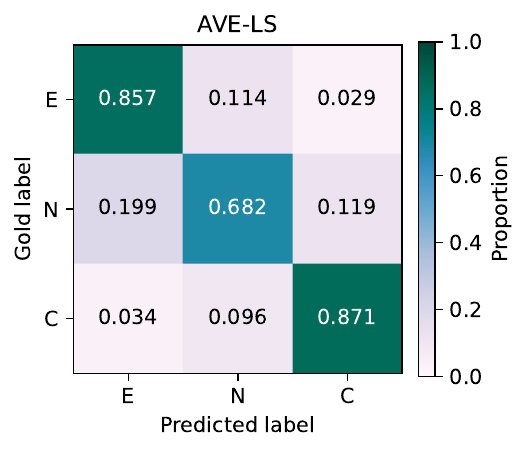}
\caption{Row-normalised confusion matrix for AVE-LS on the test set.}
\label{fig:cm_ls}
\end{figure}

Figures~\ref{fig:cm_best}--\ref{fig:cm_ls} give the full confusion matrices behind the per-class recall comparison in Section~\ref{sec:main-results}. Across all three methods, neutral is substantially confused with both other labels, while entailment and contradiction are rarely confused with each other. Errors therefore concentrate on neutral rather than between the two label extremes.

\section{Hypothesis Complexity}
\label{app:complexity}

\begin{figure}[H]
\centering
\includegraphics[width=0.9\linewidth]{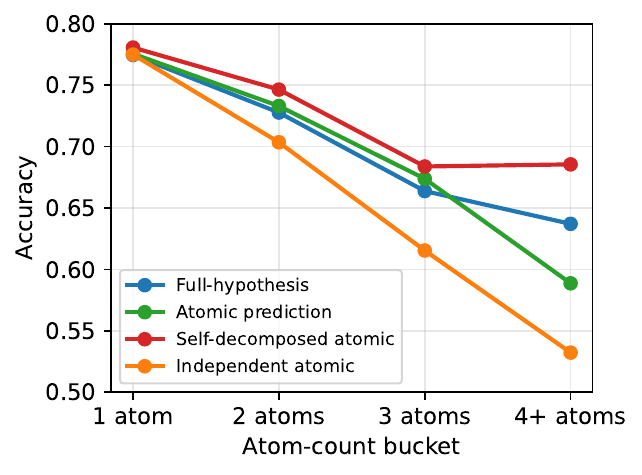}
\caption{Prediction accuracy by atom-count bucket for each method family on the test set.}
\label{fig:complexity}
\end{figure}

Figure~\ref{fig:complexity} shows accuracy by atom-count bucket for each method family. Full-hypothesis prediction declines from 0.774 at one atom to 0.637 at four or more. Independent atomic prediction falls behind the other methods from two atoms onward. Atomic and self-decomposed prediction stay ahead of full-hypothesis prediction up to three atoms, after which atomic prediction falls behind while self-decomposed prediction remains strongest.

\section{Prompts}
\label{app:prompts}
 
Figures~\ref{fig:decompose}--\ref{fig:phrase} give the full text of every prompt used. Placeholders in braces are replaced at inference time.

\begin{figure*}[p]
\refstepcounter{prompt}
\begin{tcolorbox}[promptbox, title=Prompt~\theprompt: Hypothesis decomposition,
                  label=prompt:decompose, fontupper=\scriptsize]
Break down the Claim into simple Atomic Claims.\\[3pt]
\textbf{Steps:}\\
1. Check if the Claim is simple. If yes, return it as is.\\
2. If complex, identify key components and break into distinct Atomic Claims.\\
3. Ensure Atomic Claims cover the full meaning without redundancy.\\
4. Preserve named entities and replace unclear pronouns with explicit noun phrases when needed.\\
5. Verify each fact is clear, distinct, and self-contained.\\[3pt]
\textbf{Rules:}\\
- Use only information explicitly stated in the Claim.\\
- Do not add ages, locations, motivations, intentions, causes, scene details, or implications unless explicitly stated.\\
- Keep subject, main verb, object/location together if they form one event.\\
- Do not create overlapping or duplicate Atomic Claims.\\
- Do not split a Claim into one general atom and one more specific atom that repeats the same event.\\
- Keep a clause whole if splitting it would distort the meaning.\\
- Use the fewest Atomic Claims needed.\\
- Output only the Atomic Claims inside \texttt{<answer>} \ldots \texttt{</answer>}.\\
- Do not add explanation outside \texttt{<answer>} tags.
\end{tcolorbox}
\caption{Prompt used for hypothesis decomposition.}
\label{fig:decompose}
\end{figure*}

\begin{figure*}[p]
\refstepcounter{prompt}
\begin{tcolorbox}[promptbox, title={Prompt~\theprompt: Full-hypothesis, simple prompt},
                  label=prompt:fh_simple, fontupper=\scriptsize]
Look at the image and evaluate the statement.\\[3pt]
\textbf{Statement:} \texttt{\{hypothesis\}}\\
Choose exactly one label from: \texttt{entailment}, \texttt{neutral}, \texttt{contradiction}.\\
Also, give a short explanation based only on what is visible in the image.\\[3pt]
\textbf{Return only JSON:} \texttt{\{"label": "...", "reason": "..."\}}
\end{tcolorbox}
 
\refstepcounter{prompt}
\begin{tcolorbox}[promptbox, title={Prompt~\theprompt: Full-hypothesis, structured prompt},
                  label=prompt:fh_struct, fontupper=\scriptsize]
Look at the image and evaluate the statement.\\[3pt]
\textbf{Statement:} \texttt{\{hypothesis\}}\\
First identify only the visible evidence relevant to the statement.\\
Then explain whether that evidence supports the statement, contradicts it, or is insufficient.\\
Choose exactly one label from: \texttt{entailment}, \texttt{neutral}, \texttt{contradiction}.\\[3pt]
\textbf{Return only JSON:} \texttt{\{"label": "...", "visual\_evidence": "...", "reasoning": "..."\}}
\end{tcolorbox}
\caption{Full-hypothesis prediction prompts: simple prompt (top) and structured prompt (bottom).}
\label{fig:fullhyp}
\end{figure*}

\begin{figure*}[p]
\refstepcounter{prompt}
\begin{tcolorbox}[promptbox, title={Prompt~\theprompt: Atomic, simple prompt},
                  label=prompt:at_simple, fontupper=\scriptsize]
Look at the image and perform visual entailment using the atomic facts below.\\
The atomic facts are connected parts of one complete hypothesis, so use them together.\\[3pt]
\textbf{Atomic facts:} \texttt{\{atomic\_facts\}}\\
Choose exactly one final label from: \texttt{entailment}, \texttt{neutral}, \texttt{contradiction}.\\
Also give a short explanation based only on what is visible in the image.\\[3pt]
\textbf{Return only JSON:} \texttt{\{"label": "...", "explanation": "..."\}}
\end{tcolorbox}
 
\refstepcounter{prompt}
\begin{tcolorbox}[promptbox, title={Prompt~\theprompt: Atomic, structured prompt},
                  label=prompt:at_struct, fontupper=\scriptsize]
Look at the image and perform visual entailment using the atomic facts below.\\
The atomic facts are connected parts of one complete hypothesis, so do not treat
them as fully independent. Use them together to decide whether the complete
meaning formed by these facts is entailed, neutral, or contradicted by the image.\\[3pt]
\textbf{Atomic facts:} \texttt{\{atomic\_facts\}}\\[3pt]
\textbf{Decision rules:}\\
- Choose \texttt{entailment} if the image clearly supports all important atomic facts together.\\
- Choose \texttt{contradiction} if at least one important atomic fact clearly conflicts with the image.\\
- Choose \texttt{neutral} if the image does not provide enough evidence for one or more important atomic facts, and nothing clearly conflicts.\\[3pt]
For each atom, briefly state whether it is \texttt{entailment}, \texttt{neutral},
or \texttt{contradiction}. Then explain how the atoms combine into one complete
hypothesis. Choose exactly one final label.\\[3pt]
\textbf{Return only JSON:} \texttt{\{"atom\_observations": [\{"atom": "...", "label": "...", "reason": "..."\}], "bridge\_reasoning": "...", "label": "..."\}}
\end{tcolorbox}
\caption{Atomic prediction prompts: simple prompt (top) and structured prompt (bottom).}
\label{fig:atomic}
\end{figure*}

\begin{figure*}[p]
\refstepcounter{prompt}
\begin{tcolorbox}[promptbox, title={Prompt~\theprompt: Self-decomposed atomic, simple prompt},
                  label=prompt:sd_simple, fontupper=\scriptsize]
Look at the image and perform visual entailment.\\[3pt]
\textbf{Hypothesis:} \texttt{\{hypothesis\}}\\
First, break the hypothesis into small atomic claims that can be checked against the image.\\
Then label each atomic claim \texttt{entailment}, \texttt{neutral}, \texttt{contradiction}.\\
Use only what is visible in the image.\\[3pt]
\textbf{Label meanings:}\\
- \texttt{entailment}: the image clearly supports the claim.\\
- \texttt{neutral}: the image does not provide enough evidence, and nothing clearly conflicts.\\
- \texttt{contradiction}: the image clearly shows an incompatible alternative.\\[3pt]
\textbf{Return only JSON:} \texttt{\{"decomposed\_atoms": [\{"atom": "...", "label": "...", "reason": "..."\}], "explanation": "...", "label": "..."\}}
\end{tcolorbox}

\refstepcounter{prompt}
\begin{tcolorbox}[promptbox, title={Prompt~\theprompt: Self-decomposed atomic, structured prompt},
                  label=prompt:sd_struct, fontupper=\scriptsize]
Look at the image and perform visual entailment.\\[3pt]
\textbf{Hypothesis:} \texttt{\{hypothesis\}}\\
Use the following steps:\\[3pt]
\textbf{Step 1:} Decompose the hypothesis into small atomic claims. Each atomic
claim should preserve the original meaning and should be directly checkable
against the image. Do not add new claims that are not in the hypothesis.\\[3pt]
\textbf{Step 2:} For each atomic claim, choose exactly one label from:
entailment, neutral, contradiction. Use only visible evidence from the image. Do
not infer hidden intentions, future events, causes, or background facts.\\[3pt]
\textbf{Step 3:} Aggregate mechanically:\\
- If any important atomic claim is contradiction, final label is contradiction.\\
- Else if all important atomic claims are entailment, final label is entailment.\\
- Else final label is neutral.\\[3pt]
\textbf{Return only JSON:} \texttt{\{"decomposed\_atoms": [\{"atom": "...", "label": "...", "reason": "..."\}], "bridge\_reasoning": "...", "label": "..."\}}
\end{tcolorbox}
\caption{Self-decomposed atomic prediction prompts: simple prompt (top) and structured prompt (bottom).}
\label{fig:selfdecomp}
\end{figure*}

\begin{figure*}[p]
\refstepcounter{prompt}
\begin{tcolorbox}[promptbox, title=Prompt~\theprompt: Independent atomic,
                  label=prompt:ind, fontupper=\scriptsize]
The simple and structured full-hypothesis prompts (Figure~\ref{fig:fullhyp}) are
reused, with \texttt{\{hypothesis\}} replaced by a single \texttt{\{atomic\_fact\}}.
Each fact is judged in a separate call.
\end{tcolorbox}
\caption{Independent atomic prediction. Reuses the full-hypothesis prompts
(Figure~\ref{fig:fullhyp}) applied to a single atomic fact per call.}
\label{fig:ind}
\end{figure*}

\begin{figure*}[p]
\refstepcounter{prompt}
\begin{tcolorbox}[promptbox, title=Prompt~\theprompt: Grounding phrase extraction,
                  label=prompt:phrase, fontupper=\scriptsize]
You convert one piece of visual entailment evidence into a short phrase for
visual grounding.\\[3pt]
\textbf{Return only JSON:} \texttt{\{"can\_ground": true or false, "phrase": "short visual phrase" or "NONE", "match\_phrase": "main visible entity" or "NONE"\}}\\[3pt]
\textbf{Definitions:} \texttt{phrase} is sent to the grounding model;
\texttt{match\_phrase} is the main visible entity used later for matching
Flickr30k Entities annotations.\\[3pt]
\textbf{Rules:}\\
1. If the final label is \texttt{entailment}, choose visible evidence that supports the atom.\\
2. If the final label is \texttt{contradiction}, choose what is actually visible in the image that contradicts the atom.\\
3. For \texttt{contradiction}, do not describe the false hypothesis claim.\\
4. If the evidence is only about absence or a missing object, return \texttt{NONE}.\\
5. Keep the phrase concrete and visible.\\
6. The phrase should be at most six words.\\
7. The \texttt{match\_phrase} should be shorter than the phrase and should name the main visible entity.\\
8. Do not explain.\\[3pt]
Three worked examples follow in the prompt, covering entailment, contradiction,
and an absence-only case that returns \texttt{NONE}.
\end{tcolorbox}
\caption{Grounding phrase-extraction prompt.}
\label{fig:phrase}
\end{figure*}

\end{document}